\documentclass[twocolumn]{article}

\usepackage[utf8]{inputenc}
\usepackage{authblk}
\usepackage{dcolumn}
\usepackage{float}
\usepackage{xcolor}
\usepackage{graphicx} 
\usepackage{subfigure}
\usepackage{amsmath}
\usepackage{placeins}
\usepackage{derivative}
\usepackage{fancyhdr} 
\usepackage{geometry} 
\usepackage{setspace}
\usepackage{nomencl}
\usepackage{array}
\usepackage{amssymb}
\usepackage{pdfpages}
\usepackage{listings}

\usepackage{multicol}  
\usepackage{titlesec}

\title{A Deep-Learning Method Using Auto-encoder and Generative Adversarial Network for Anomaly Detection on Ancient Stone Stele Surfaces}

\author{Yikun Liu$^{\text{1}}$, Yuning Wang$^{\text{2}}$, Cheng Liu$^{\text{1}}$}
\date{
1: School of Cultural Heritage, Northwest University, Xi'an, China \\
2: FLOW, Engineering Mechanics, KTH Royal Institute of Technology, SE-100 44 Stockholm, Sweden.
}

\begin{document}

\twocolumn[
  \begin{@twocolumnfalse}
    \maketitle
    \begin{abstract}Accurate detection of natural deterioration and man-made damage on the surfaces of ancient stele in the first instance is essential for their preventive conservation. Existing methods for cultural heritage preservation are not able to achieve this goal perfectly due to the difficulty of balancing accuracy, efficiency, timeliness, and cost. This paper presents a deep-learning method to automatically detect above mentioned emergencies on ancient stone stele in real time, employing autoencoder (AE) and generative adversarial network (GAN). The proposed method overcomes the limitations of existing methods by requiring no extensive anomaly samples while enabling comprehensive detection of unpredictable anomalies. the method includes stages of monitoring, data acquisition, pre-processing, model structuring, and post-processing. Taking the Longmen Grottoes' stone steles as a case study, an unsupervised learning model based on AE and GAN architectures is proposed and validated with a reconstruction accuracy of 99.74\%. The method's evaluation revealed the proficient detection of seven artificially designed anomalies and demonstrated precision and reliability without false alarms. This research provides novel ideas and possibilities for the application of deep learning in the field of cultural heritage.
    
    \end{abstract}
  \end{@twocolumnfalse}
]
\maketitle
\section{Introduction}



Ancient stone steles, which are found at almost every major heritage site in China, represent a significant part of the cultural heritage. 
However, they are prone to various environmental degradation and man-made damage, urgently needing effective preventive conservation.

Timely detection of these risks is considered as the prerequisite of their preventive conservation.
Commonly employed strategies to detect risks for these outdoor relics include regular investigation by professional conservators, and daily patrols by heritage managers or safety officers. Nevertheless, these methods have their limitations. The former requires a wealth of specialized resources and doesn't provide immediate feedback, while the latter, despite its regularity, often overlooks slowly evolving deterioration due to a lack of specialization.

Over recent years, deep learning techniques have been increasingly applied in the field of cultural heritage\cite{mishra2021machine}. Although certain studies have ventured into automated deterioration recognition, these solutions are not sufficiently adapted to meet the needs inherent in preventive conservation. 
Current research primarily employs supervised learning based on deep neural networks(DNN) to achieve classification~\cite{hatirDeepLearningbasedWeathering2020},~\cite{meklati2023surface},~\cite{cao2020ancient}, object detection~\cite{mishraArtificialIntelligencebasedVisual2022},~\cite{zou2019cnn}, and semantic segmentation~\cite{hatirDeepLearningMethod2021},~\cite{liuSemanticSegmentationPhotogrammetry2022} of various forms of deterioration and damage. 
However, these methods entail an over-reliance on a large number of high-quality labeled samples, a requirement that greatly limits their application.
In the cultural heritage domain, samples related to deterioration and damages are terribly difficult to obtain sufficiently for supervised deep learning.
This is further complicated by the diversity of the deterioration and damages, which further escalates the total requirement of samples. 
For example, patterns of man-made damages such as doodle and carving are visually unpredictable, making it impossible to ensure that they are learned by the model. 
This directly hinders the possibility of these deep learning methods to be used for discovering emerging risks.
In summary, novel approaches are needed to realize the potential of deep-learning techniques in preventive conservation of cultural heritage.

\section{Research aim} 
The research aims at develop a deep-learning method based on auto-encoder(AE) and generative adversarial network(GAN) for real-time automatic detection towards deterioration and damages (hereinafter collectively referred to as 'anomaly') appearing on the surface of ancient stone steles.

The main technical features this method aims to implement are as follows: eliminates the extensive requirement for anomaly samples during modeling and remains sensitivity towards unpredictable anomalies during detection.

To this end, firstly an HD camera is installed to monitor the surface of the stone stele and to collect normal samples as training data. Then a AE/GAN-based neural network is built to reconstruct the image features of these normal samples. Subsequently, Post-processing methods are designed to detect and locate anomalies by comparing the reconstruction difference between abnormal and normal samples. Finally, the effectiveness of the method was verified by a series of simulated test images containing various types of anomalies and ambient lighting conditions.

\section{Study area}

The Longmen Grottoes, situated on both sides of the Yi River south of Luoyang in China, serve as the study area for this research, focusing on an open-air stone stele at the Fengxian Temple.
In addition to tens of thousands of Buddha statues and more than 60 stupas, there are about 2,800 steles with inscriptions at the heritage ~\cite{UNESCO}.

\begin{figure}[!ht]
    \centering
    \includegraphics[width=0.47\textwidth]{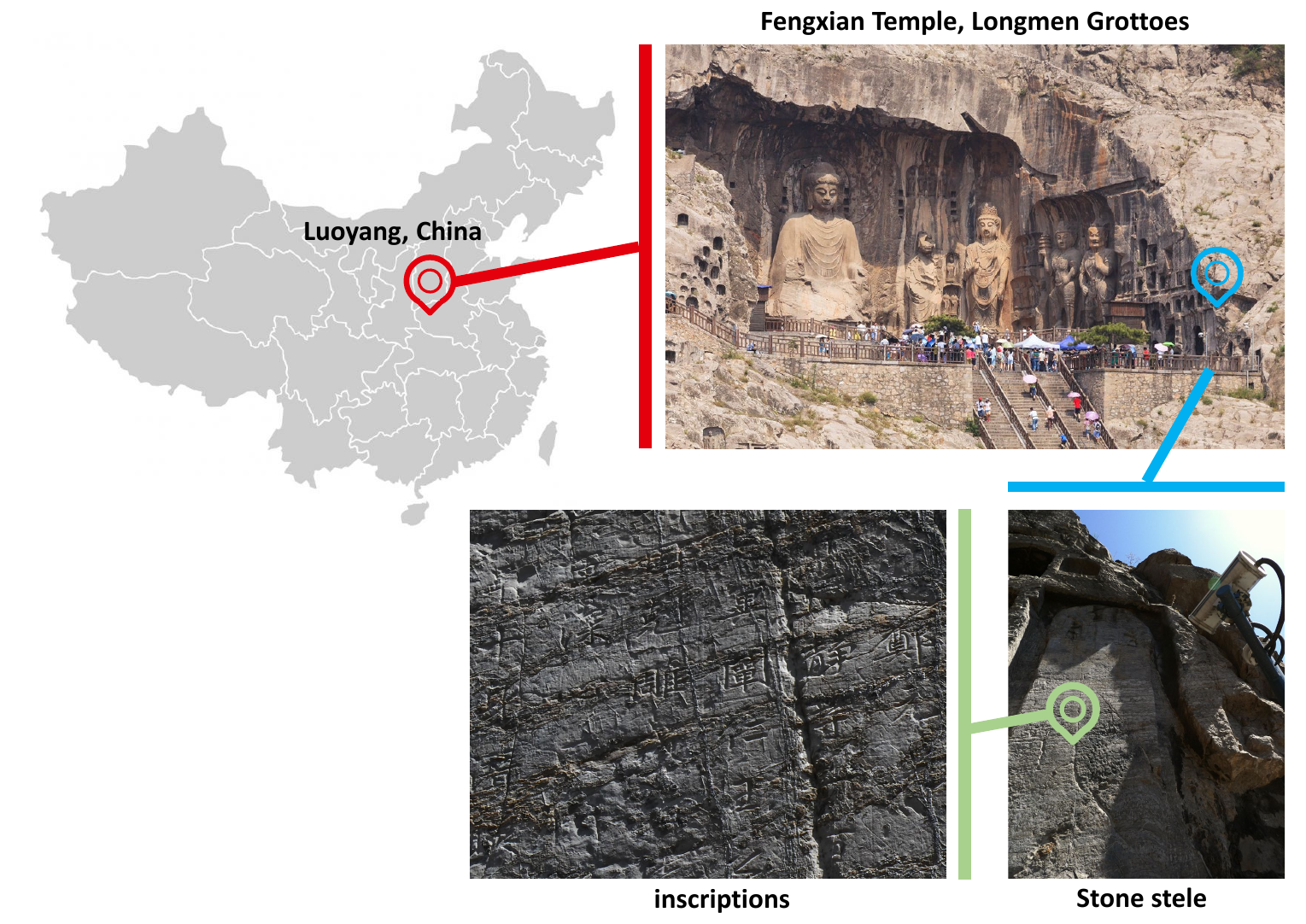}
    \caption{Location of the study area.}
    \label{fig:Study area}
\end{figure}

The region where the grotto is located has a temperate continental climate with cold arid winters, hot rainy summers and high windy spring and fall seasons, resulting in the natural deterioration of this open-air stele over a long period of time ~\cite{xu2012research},~\cite{yun2013application}.
Signs of Surface weathering are prevalent on the stele, manifested as lateral surface dissolution and extensive black deposit accumulation.
Furthermore, the presence of top-down crack indicates a potential structural safety risk that warrants continued attention. Additionally, this stele, due to its accessibility to large amount of tourists, is highly susceptible to accidental human-induced damage from the proximate interaction.

\section{Proposed methods}
\subsection{Overview of the method}


In response to the need for preventive conservation of cultural heritage, the proposed method is supposed to automatically track the emergence of both natural deterioration and man-made damage on the stone stele surface. Specifically, we expect that these anomalies should be accurately located and rapidly reported to the conservators at the first sign of occurrence, so that timely intervention can be made before more serious damage occurs.

\begin{figure*}[!ht]
    \centering
    \includegraphics[width=\textwidth]{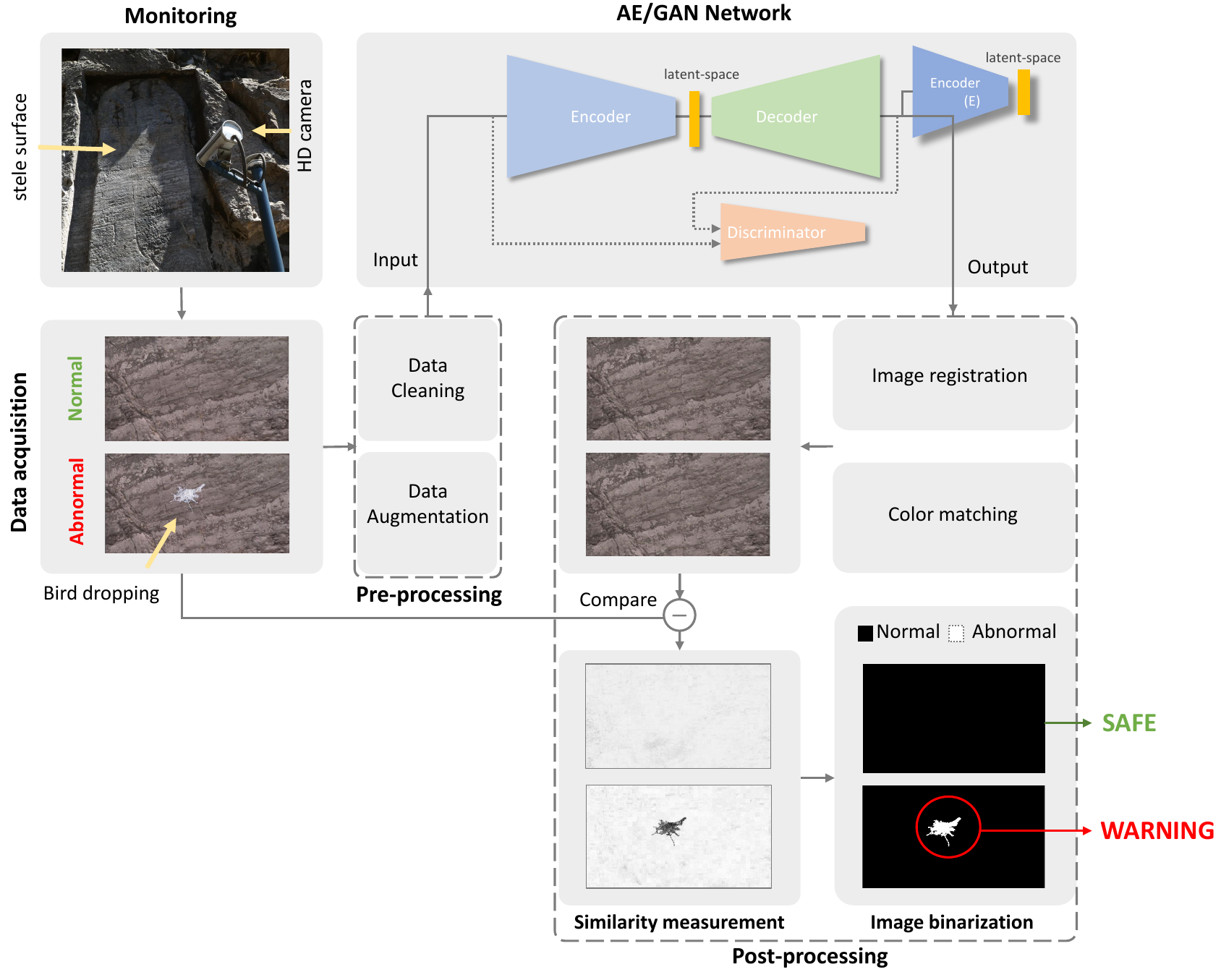}
    \caption{Flowchart depicting the comprehensive process for anomaly detection on ancient stone steles. An artificial designed anomaly image with bird dropping is used as an example to show how the proposed method works.}
    \label{fig:Overview}
\end{figure*}

The scheme of our proposed method is as follows (Fig.~\ref{fig:Overview}). Initially, a fixed camera serves as the monitoring instrument, capturing images of the stele surface in its normal state. These images then form the training dataset for a model we developed based on Auto-encoder and Generative Adversarial Network (GAN) frameworks. This trained model is capable of reconstructing images of the stele surface in its normal state. When an image of the stele surface containing anomalies is fed to the model in the future, since they are never learned by the model, the reconstructed image and the input image differ significantly where the anomalies are located. Finally, by choosing appropriate methods to measure this difference, it can be determined whether the stele is in an emergency situation, and the location of the anomaly will be segmented and presented as a binary image.

\subsection{Data acquisition}
 The lower center area of the stele has been chosen as monitoring area, considering that most of the natural deterioration and man-made damage risks are concentrated here.   
    
A high-definition camera fixed to the side of the stele is used as the monitoring device. This camera was programmed to capture the surface of the stele regularly from an identical angle, thereby ensuring the continuity and comparability of the collected images. Each of these images bore a resolution of 3840 × 2748 pixels, featuring both vertical and horizontal pixel densities of 96 dpi, and recorded in 24-bit RGB three-channel color. This ensures that the majority of macro-scale surface changes can be accurately documented. The camera's shooting frequency is set at two images per day, one in the morning and one in the afternoon. This is a compromise between maximizing data diversity to train the model and minimizing the storage footprint when collecting high-resolution images over long periods of time.

\subsection{Pre-processing}
Pre-processing of the collected images is an integral step to enhance the quality of our training data.

Firstly, noisy data such as severe overexposure and foreign object occlusion are removed by manual inspection. This approach contributes to improving model accuracy. However, it  has to be mentioned that when such noise reoccurs in the future, they could potentially trigger false alarms.

Subsequently, the original photographs are cropped and partitioned into six equal-sized regions. Each of these regions is resized to a 640x480 resolution, maintaining the RGB three-channel color, thus preserving the color information vital for anomaly detection. Therefore, the input size for the proposed model is fixed at 640x480x3. For each distinct region, an independent model will be trained. When deployed, these six models will operate in parallel.The above procedure are taken to prevent potential negative impacts such as model training difficulties and overfitting, which might be brought about by overly high resolution. For illustrative purposes, this study only showcases one selected area.

Finally, we implement data augmentation by adjusting the images' white balance and exposure. This adjustment is conceived to simulate the variations in photographic parameters under different lighting conditions. For each image, the exposure value and the white balance are altered randomly, once each, within a range of +/- 1EV and +/- 1000K, respectively. As a result, the volume of training data is amplified by a factor of two. Notably, given the fixed angle and focal length from which all photographs were taken, certain common data augmentation techniques such as rotation, flipping, and zooming were consequently disregarded in our approach.

\subsection{Proposed neural network}
\begin{figure}[!ht]
    \centering
    \includegraphics[width=0.4\textwidth]{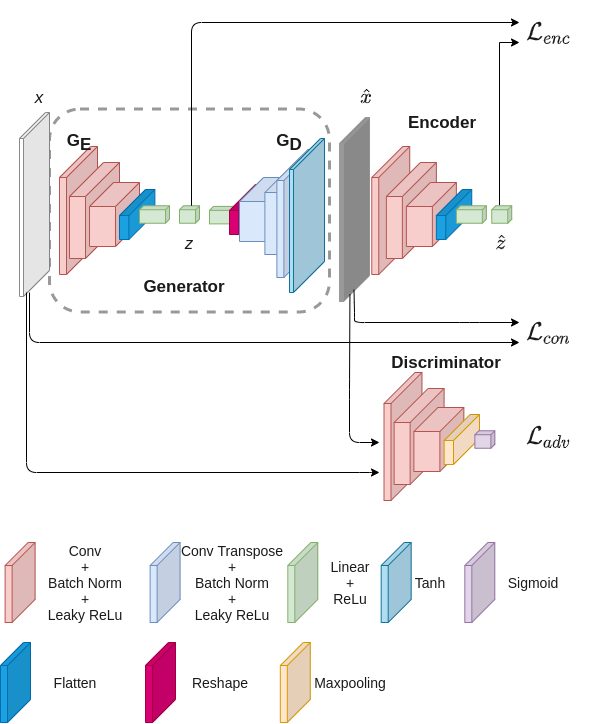}
    \caption{Schematic pipeline of the proposed GANomaly architecture. Colored cubes denote different types of layer which are implemented for compositing each sub-networks. The generator is annotated via a black rectangle. Dashed lines denote features obtained through hypothesis of sub-network while arrows point to corresponding loss functions.}
    \label{fig:GAN_pipeline}
\end{figure}

We proposed an modified, unsupervised-learning model based on \textit{GANomaly}~\cite{GANnomaly_Org}, which is a branch of Generate Adversarial Network (GAN), specifically designed for anomaly detection. The architecture, as shown in Fig.~\ref{fig:GAN_pipeline}, comprises a generator (${G}$), an Encoder (${E}$) and a Discriminator (${D}$). 

The generator $G$, following the bow-tie architecture of autoencoder~\cite{hinton2006reducing}, is designed for learning latent-space representation $z$. The input data from the reference space $x \in \mathcal{X}$ will be reduction of encoder $G_E$ and we employ a decoder $G_D$ to project the representation to the reference space. The second sub-network is the discriminator $D$ which is used to classify input $x$ and reconstructed image $\hat{x}$. This sub-network is the standard discrimator network introduced in DCGAN~\cite{DCGAN}. The third sub-model is the encoder network $E$ that compress the reconstructed image $\hat{x}$. It owns the same architecture of generator encoder $G_E$ while holding different parameters. Note that the encoder $E$ is not taken into the training process.  $E$ compress $\hat{x}$ to derive its latent-space features vector $\hat{z}$ which has identical dimension as features vector $z$. We take the $l_2$ norm error between $z$ and $\hat{z}$ is a part of loss function. This sub-network is the unique parts of GANonmaly~\cite{GANnomaly_Org}, which used for stablise the latent space and imporve the reconstruction quality. For encoders and decoders in the architecture, we adopt convolution neural networks (CNNs) for learning hierarchical representations from the original image domain, aiming at extract the anomaly features from the abnormal image.

The input and output of the model, are the image of normal stele $x \in \mathbb{R} ^{w\times h \times 3}$, where $w$ and $h$ denotes the width and height of the image, respectively. The input is first propagated to its encoder $G_E$ where data is compressed to a vector $z\in \mathbb{R} ^{d}$ by three blocks comprises CNN layers which used for learning hierarchical representation  and linear layers. $z$ is known as the latent-space features of input data and hypothesised to obtain the lowest-dimension representation of $x$ via $z = G_E(x)$. Subsequently, the latent space representation $z$ is fed to decoder of generator $G_D$, implementing transposed CNN layers for project the representation back to the reference space. The function of decoder is to upscale the vector $z$ to reconstruct the image $x$ as $\hat{x} \in \mathbb{R} ^{w\times h \times 3} $ via $\hat{x} = G_D(z)$. We propose contextural loss to update the trainable parameters in generator by: 
\begin{equation}
    {\mathcal{L}_{con}} = {\mathbb{E}_{x \sim {p_{\mathbf{X}}}}}{\parallel}{x-G(x)}{\parallel}_1
    \label{eq:con_loss}
\end{equation}

The reconstructed image, so-called \textit{fake} output in GAN model, is fed to the discriminator as a supervision to penalise the generator. In another word, the generator is trained for "fool" the discriminator, thus we introduce adversarial loss to penalise the model during training:
\begin{equation}
    {\mathcal{L}_{adv}} = {\mathbb{E}_{x \sim {p_{\mathbf{X}}}}}{\parallel}{\mathnormal{f}(x)} - {\mathbb{E}_{x \sim {p_{\mathbf{X}}}}}{\mathnormal{f}(G(x))}{\parallel}_2
    \label{eq:adv_loss}
\end{equation}

Additionally, we employ an extra encoder loss ${\mathcal{L}_{enc}}$ to minimize the distance between the latent features vector of input ($z = {G_E}(x)$) and the encoded features of the generated image ($\hat{z} = E(\hat{x})$). Through this optimization, the generator is able to learn how to encode features of the generated image for normal samples. ${\mathcal{L}_{enc}}$ is defined as:
\begin{equation}
    {\mathcal{L}_{enc}} = {\mathbb{E}_{x \sim {p_{\mathbf{X}}}}}{\parallel}{G_E(x)-E(G(x))}{\parallel}_2
    \label{eq:enc_loss}
\end{equation}

Overall, our total loss function for generator becomes the following: 
\begin{equation}
    {\mathcal{L}} = {\omega_{adv}}{\mathcal{L}_{adv}} + {\omega_{con}}{\mathcal{L}_{con}} + {\omega_{enc}}{\mathcal{L}_{enc}}
    \label{eq: total_loss}
\end{equation}
\noindent where ${\omega_{adv}}$, ${\omega_{con}}$ and ${\omega_{enc}}$ are the weight of corresponding loss object which adjusts the impact of individual losses to the overall loss function.
In the present study, we adopt ${\omega_{adv}}=1$, ${\omega_{con}}=40$ and ${\omega_{enc}}=1$ for each weighting respectively.

\subsection{Post-processing}


%
After the completion of model training and deployment, the raw output cannot be immediately used for anomaly detection. Post-processing is designed to quantify the difference between the input and reconstructed images, and subsequently facilitate binary classification in the form of segmentation normal and abnormal. The post-processing steps in this study include: image registering, color matching, similarity measurement, binarization. 
\subsubsection{Image registration}
The reconstructed image is not perfectly aligned with the input image. The complex texture of the stele surface means any misalignment can introduce substantial noise into the subsequent similarity measurements, thereby undermining the accuracy of anomaly detection. Hence, to address this issue, we incorporate image registration, a technique that uses image processing algorithms to spatially align multiple images. Specifically, we adjust the texture of the reconstructed image with reference to each input image, thus ensuring that they align more precisely and are better prepared for subsequent processing stages.

To this end, we employ the algorithm so-called Speeded-Up Robust Features (SURF), a type of feature-based registration, for this task. SURF operates by autonomously identifying key feature points and then accomplishing registration by matching them. This method is particularly appropriate for our study as it is well-suited in cases where there are discrepancies in brightness and contrast, meeting the need for robustness under this study. The image registration in this study can be simplified as:

\begin{equation}
    \hat{X}' = \text{SURF}(X, \hat{X})
    \label{eq: image registration}
\end{equation}

Where $X$ and $\hat{X}$ represent the input image and the reconstructed image respectively. $\hat{X}'$ is the reconstructed image after registration. $\text{SURF}$ is the function representing the SURF algorithm. 

\subsubsection{Color matching}
Color difference between the input and reconstructed images can also confound the accuracy of the subsequent similarity measurements in our study. we implemented color matching to make the reconstructed image as closely aligned as possible to the input image in terms of exposure and color temperature, without altering the texture of the image.

Specifically, we begin by normalizing the pixel values of each of the three channels in the reconstructed image, a process that involved subtracting the mean and dividing by the standard deviation. Following normalization, we referenced the input image to rescale and recenter these values. This involved multiplying the normalized values by the standard deviation of the input image, and then adding the mean of the input image. As a result, the distribution of pixel values in the reconstructed image came into alignment with those of the corresponding channels in the input image. This process is defined as:
\begin{equation}
    \hat{X}_{ijk}'' = \frac{{\hat{X}_{ijk}' - \mu_{\hat{X}_k}}}{{\sigma_{\hat{X}_k}}} \cdot \sigma_{X_k} + \mu_{X_k},
    \label{eq: color matching}
\end{equation}

\noindent where $\hat{X}_{ijk}''$ is the pixel value in a specific channel of reconstructed image after color matching, $i$, $j$ are the pixel coordinates and $k$ is the channel index. $\sigma_{X_k}$, $\sigma_{\hat{X}_k}$, $\mu_{X_k}$ and $\mu_{\hat{X}_k}$ are the average and standard deviation of all pixels respectively in a specific channel.

\subsubsection{Similarity measurement}

Given the necessity to localize anomalies, similarity matrix rather than mere similarity value is employed for similarity measurement. 
Calculating the reconstruction error of each pixel directly by matrix subtraction(MS) is the most straightforward and efficient method. 
Specifically, we first subtract the reconstructed image pixel by pixel from the input image and then decenter the result by the median of each RGB channel. Finally we linearly stack the three channels to obtain the similarity matrix.
The decentering calculation is performed to make the reconstruction error small enough in regions without anomalies, which in turn allows a better distinction between normal and abnormal.
The operation for each pixel is defined as:
\begin{equation}
\begin{aligned}
    & \text{MS}(X, \hat{X}'') = \\
    \sum_{k} & \left| X_{ijk} - \hat{X}_{ijk}'' - \text{med}_{k}(X_{ijk} - \hat{X}_{ijk}'') \right|
    \label{eq: matrix subtraction}
\end{aligned}
\end{equation}
\noindent where $\text{MS}$ represents the matrix subtraction method, and $\text{med}_k$ stands for median calculation on the $k$ channels.

However, After a series of experimental tests, we observed that this method exhibited insufficient sensitivity to differences in image structure, thus resulting in compromised detection results.
This could be attributed to the fact that the computation in this method occurs at an individual pixel level, without taking into account the correlation between pixels.

To ensure a comprehensive coverage of various anomalies in similarity measurements, we sought to complement the matrix subtraction method with the introduction of the structural similarity index (SSIM) ~\cite{wang2004image}. 
The SSIM method is designed to mimic human perception by focusing on brightness, contrast, and structural similarities between images. 
In our study, we first convert input and reconstructed images to grayscale by linearly stacking the three color channels to comply with the SSIM requirements for input.
Then we compare the difference between patches rather than individual pixels of the input and reconstructed images with SSIM. 
In the subsequently obtained similarity matrix, each pixel corresponds to the similarity value of the patch pair which takes that pixel as a vertex.
The SSIM for each patch are defined as:

\begin{equation}
\begin{aligned}
    \text{SSIM}(X, \hat{X}'') = \frac{{2\mu_{X_g}\mu_{\hat{X}_g''} + C1}}{{\mu_{X_g}^2 + \mu_{\hat{X}_g''}^2 + C1}} \\
    \cdot \frac{{2\sigma_{X_g\hat{X}_g''} + C2}}{{\sigma_X^2 + \sigma_{\hat{X}_g''}^2 + C2}},
    \label{eq: SSIM}
\end{aligned}
\end{equation}
\noindent where ${X_g}$ and ${\hat{X}_g''}$ are the greyscale version of $X$ and $\hat{X}''$. The corresponding mean, standard deviation and mutual covariance are denoted by $\mu$, $\sigma$ and $\sigma^2$, respectively. 
$C1$, $C2$, and $C3$ are the regularization constants of brightness, contrast and structural terms, respectively.

\subsubsection{Binarization}


Utilizing an appropriate detection threshold, the similarity value each pixel represents within the matrix obtained from the prior section is categorized into one of two classes—normal or abnormal.
The above process consequently yields a binary image. This image provides an outline of anomalies present on the input image in the form of image segmentation, enabling the detection and localization of the anomalies.
However, it must be noted that the choice of detection threshold significantly influences the quality of detection. Setting the threshold too low could label an excessive number of pixels as anomalies, potentially causing false alarms due to noise, whereas setting it too high could lead to the opposite outcome.

Taking these factors into account, this study empirically selects a moderate detection threshold and further applies a denoising process to the binary image to mitigate potential noise.
The determination of the optimal detection threshold is not discussed in this paper.
Considering that the most notable difference between noise and genuine anomalies lies in the area size, the denoising process is implemented as follows: an area value is determined, and pixel regions in the binarized image smaller than this value are considered noise and subsequently removed. 

The binarization and noise reduction processes are simultaneously applied to the two similarity matrices obtained from both matrix subtraction (MS) and structural similarity index (SSIM) calculations to generate two binarized images. By taking the union of these two images on a pixel-by-pixel basis, the resultant image serves as the final anomaly detection output.

\section{Experimental Results}

\subsection{Training details}
In the present study, we acquire high-definition normal images of stele surface collected over a period of six-month as initial data. After eliminating instances when the camera malfunctioned or when foreign objects obstructed the view, a total of 283 usable images remained. Thanks to image augmentations, the scale of the data is enhanced to 849. We use 840 images as training dataset while the rest 9 were utilized to create artificially anomalous conditions, serving as test images for method assessment.


The training details are summarized in Tab~.\ref{tab:train_summary}. Once we have trained a model for a region, we first evaluate the reconstruction accuracy by computing the relative $l_2$ norm error as:
\begin{equation}
    E_{\rm rec} = \frac{|| y - \Tilde{y}||_2 } {|| y ||_2} \times100\%,
\end{equation}
\noindent where $y$ denotes the ground truth and $\Tilde{y}$ denotes the prediction, respectively. In the Tab~.\ref{tab:train_summary}, we also report the reconstruction accuracy averaged over all 6 regions, which is denoted in red. Our proposed model achieves 99.74\% accuracy on whole test dataset, indicating the promising reconstruction performance of the models, which paves the way for further post-processing.

\begin{table}[!ht]
\centering
\resizebox{\columnwidth}{!}{
    \begin{tabular}{ccc}
        \hline
        \textbf{Optimizer}  & \textbf{Weight Decay} & \textbf{Learning Rate}                      \\ \hline
        Adam                & $1 \times 10^{-7}$              & $1 \times 10^{-3}$                          \\ \hline
        \textbf{Batch Size} & \textbf{No.Epoch}                  & \textcolor{red}
        {${\boldsymbol{E_{\rm rec}}}$ (\%)} \\ \hline
        16                  & 1200                            & \textcolor{red}{0.26}                                       \\ \hline
    \end{tabular}%
}
\caption{Summary of the training setup for each model and the achieved reconstruction error averaged over all segments.}
\label{tab:train_summary}
\end{table}

\subsection{Method evaluation}



In addition to reconstruction accuracy, it is also crucial to evaluate the performance of the proposed method to effectively detect anomalies.
To this end, an evaluation plan was developed using the 9 spare images from the original dataset.
These images were artificially edited to simulate a range of potential future deterioration and damages patterns that might occur on the stele surface.
By artificially creating these anomalies, we could ensure representativeness in our testing conditions. Moreover, this approach made it easier to compare results within the same surface texture, providing a more controlled evaluation environment.

The artificially introduced anomalies covered 7 categories: carving, crack, moss, doodle, salt, water stains and bird dropping.
They are designed considering the diversity both in brightness, color, and structural patterns.
From each of the nine normal images, seven artificially anomalous images were created, resulting in a total of 63 anomaly images for method evaluation.
Thus, coupled with the original 9 normal images, the evaluation dataset contains a total of 72 images.

\begin{figure*}[!ht]
    \centering
    \includegraphics[width=\textwidth]{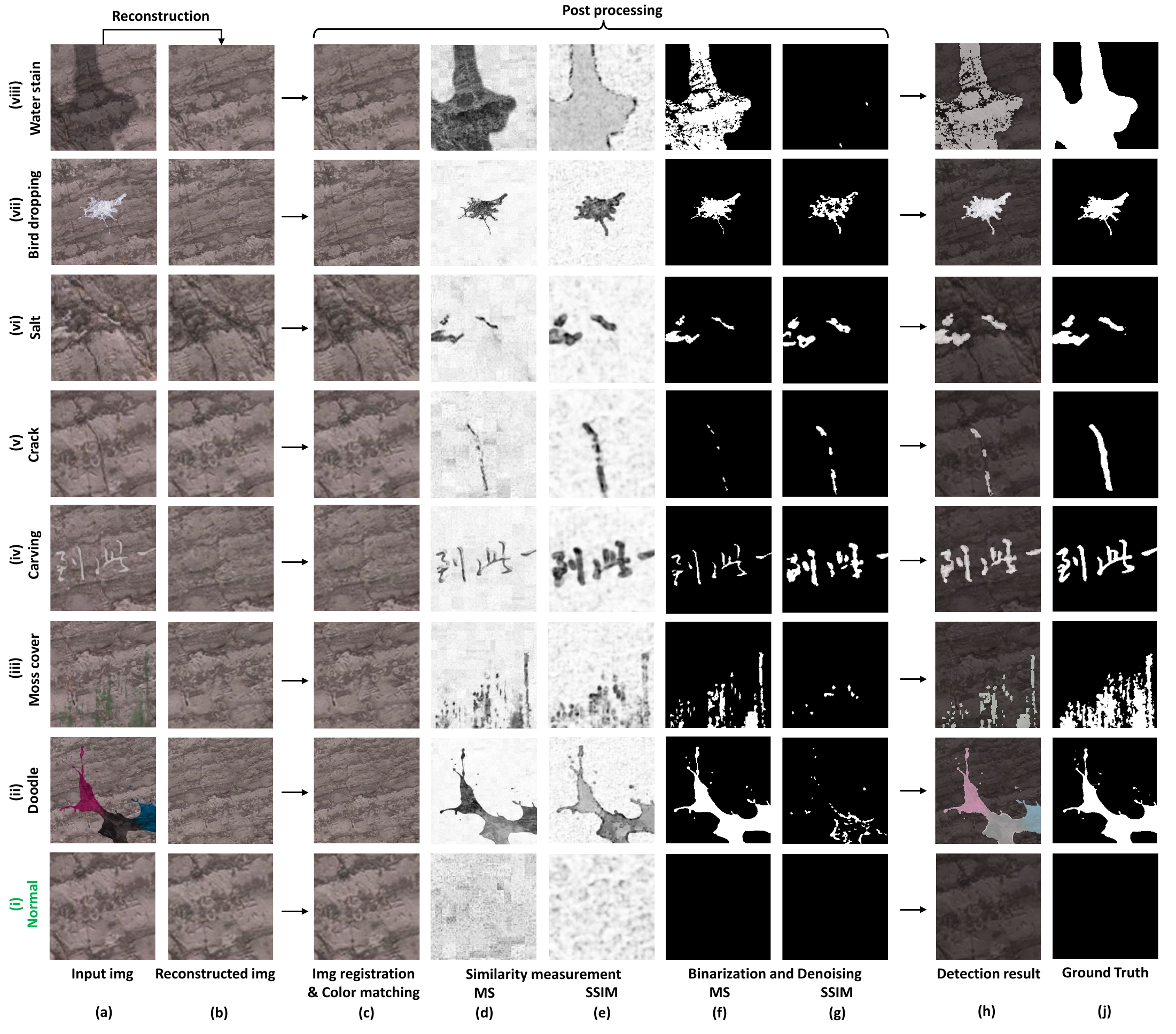}
    \caption{The experiment results obtained from testing normal and abnormal images. The types of the image are denoted at the left panel, and the processes of the proposed method are denoted at the bottom, respectively.}
    \label{fig:results}
\end{figure*}
%

The test results of the 8 images are shown in Fig.~\ref{fig:results}, including the raw output, the intermediate results of the post-processing steps, and the comparison of the final detection results with Ground Truth.

Our evaluation reveals that in the region outside the anomalies, the raw outputs are strikingly similar to the input images, indicating the model's impressive reconstruction capabilities. The result of the reconstruction of the anomalies is consistent with the original texture of the stele surface. Particularly,  in contrast to (i.b), the normal image (i.a) was almost completely reconstructed. 

Although high reconstruction accuracy makes the enhancements offered by image registration and color matching(c) virtually indiscernible to the unaided eye, it is essential not to discount these steps. The subtle improvements they bring may play a crucial role in guaranteeing the accuracy of the similarity measure. To quantify their effect on the model's reconstruction results and better evaluate their contribution, we aim to conduct a more detailed examination in future studies.

The similarity measurement results are graphically represented as heatmaps, where darker hues denote larger differences(d)(e). Clearly, the areas with anomalies showcase significantly larger reconstruction errors compared to the rest of the regions, allowing for a well-defined outlining of these anomalies. However, an unavoidable element of noise is present within the results generated by both the Matrix Subtraction(MS) and Structural Similarity Index(SSIM). Since the values of these noises are close to the anomalies, they can negatively affect the binarization process by blurring the distinction between the anomaly-induced differences and noise.

The binarization results(f)(g) stem from the empirically selected thresholds for Matrix Subtraction (MS) and Structural Similarity Index (SSIM), including  the respective detection and noise reduction thresholds. This paper doesn't delve into discussion on the selection of these thresholds.
From the results, it is evident that MS and SSIM exhibit variable proficiency in detecting different types of anomalies. Both salt(vi) and bird droppings(vii) are proficiently detected by both methods, largely due to their distinctive brightness, color, and structural differences compared to the stele surface.
On the other hand, the differences between carvings(iv) and cracks(v) and the stele surface are mainly structural, making SSIM a more effective detection method for these types of anomalies. Conversely, in the case of doodle(ii), moss cover (iii), and water stains(viii), SSIM's performance falls short due to the high structural similarity of these anomalies with the stele surface, making them nearly undetectable. While MS maintains a good detection performance in these cases.

Comparing the final detection results(h) with the ground truth(j), it is observed that after combining the detection results of MS and SSIM, our method successfully detects all 7 types of anomalies. 
There are slightly differences in the detection results for moss cover(iii) and cracks(v) when compared with GT, showing as part of the anomalies region is not detected. 
However, this discrepancy does not significantly impede the effectiveness of the detection. 
Notably, both methods excel in normal(i) detection, yielding no false alarms, demonstrating their precision and reliability.

\section{Conclusion}
The present study introduces a deep-learning method for real-time automatic detection of natural deterioration and human damage on ancient stone steles using Longmen Grottoes as an example.
Utilizing the model architecture of auto encoder (AE) and generative adversarial network (GAN), the proposed method offers the ability to eliminate the requirement for extensive anomaly samples while maintaining sensitivity towards unpredictable anomalies, which eventually addresses the limitations in existing deep learning methods for heritage deterioration recognition.

The proposed model achieves a reconstruction accuracy of 99.74\% with small architecture and dataset, which indicates the promising performance of the proposed model.
Regarding post-processing, the similarity measurement strategy combining Matrix Subtraction (MS) and Structural Similarity Index (SSIM) comprehensively covers the differences between the input and reconstructed images in terms of brightness, color, and texture structure.
By choosing appropriate thresholds, the binarization and denoising process are able to accomplish the two-class distinction between normal and abnormal well, and the results can be directly used as the detection results.

In the final method evaluation, all seven types of artificially designed anomalies were successfully detected without false alarms to normal conditions, demonstrating its exceptional precision and reliability.
Some minor discrepancies in detection, such as partial undetection of certain anomalies like moss cover and cracks, pinpoint areas for future refinement. 
The proposed method provides novel scenarios and thoughts for the application of deep learning in the field of preventive conservation, by demonstrating a tool for risk detection that is superior in both efficiency and capability. 
Future research may further explore the selection of optimal thresholds and continue to fine-tune the model for higher accuracy and a wider range of application scenarios.

\bibliographystyle{unsrt}
\bibliography{ref}


\end{document}